\documentclass[runningheads]{llncs}
\usepackage{hyperref}
\usepackage[T1]{fontenc}
\usepackage{booktabs}
\usepackage[table]{xcolor}

\usepackage{multirow}
\usepackage{graphicx,verbatim}
\usepackage{amsmath}
\usepackage{bbm}
\usepackage{tcolorbox}
\usepackage{xcolor}
\newcommand{\mytag}[1]{{\setlength{\fboxsep}{1.5pt}\colorbox{gray!20}{\texttt{#1}}}}

\begin{document}

%


\title{
Look-Closer-Then-Diagnose: Confidence-Aware Ultrasound VQA via Active Zooming
}
%

\author{Yue Zhou\inst{1, 2} \and Erxuan Wu\inst{3} \and Yikang Sun\inst{3} \and Hongjoo Lee \inst{1} \and \\ Yuan Bi\inst{1,2} \and Huixiong Xu\inst{3} \and Nassir Navab\inst{1,2}, Zhongliang Jiang\inst{4}}  

\institute{Computer Aided Medical Procedures (CAMP),\\ TU Munich, Germany \\
\and Munich Center for Machine Learning (MCML), Munich, Germany \\
\and Zhongshan Hospital, Fudan University, China
\and The University of Hong Kong, Hongkong, China}

\authorrunning{Yue Zhou et al.}
\titlerunning{Look-Closer-Then-Diagnose}

\maketitle              
\begin{abstract}
Vision-Language Models (VLMs) have significantly advanced medical visual question answering, yet their performance in ultrasound remains suboptimal. In clinical practice, sonographers explicitly focus on lesion regions to formulate reports, though diagnostic interpretations sometimes vary due to inherent subjectivity. However, existing VLMs are not explicitly structured to interactively zoom into lesions prior to diagnosis; moreover, they typically treat annotations as unbiased ground truths, failing to account for their inherent subjectivity and ambiguity. In this paper, we propose a framework specifically designed to consider the sonographer's cognitive workflow. We first introduce a structured Zoom-then-Diagnose paradigm, which replicates the interactive search process to enable lesion-focused reasoning. Furthermore, within the Group Relative Policy Optimization (GRPO) framework, we introduce an uncertainty-aware reward derived from stochastic group-wise rollouts to estimate prediction consistency as a proxy for model confidence. Together, these two components encourage the model to reinforce accurate predictions on clear cases while remaining cautious under ambiguity.
Experiments across liver, breast, and thyroid datasets show that our framework improves lesion localization by 39.3\%, demonstrating that our model has learned the ability to actively look closer and diagnose. 


\end{abstract}

\section{Introduction}
\begin{figure}[t]
  \centering
  \includegraphics[width=\textwidth]{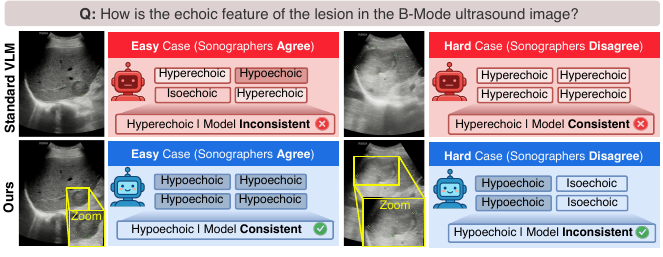}
  \caption{Our approach mimics sonographer cognitively. It employs a zoom-in mechanism to simulate sonographers' localized visualization. Moreover, it reflects their consensus: high consistency in easy cases (where sonographers agree) and appropriate ambiguity in hard cases (where sonographers disagree).}
  \label{fig:example_image}
\end{figure}

Building on general Vision-Language Models (VLMs)~\cite{bai2025qwen2,chen2024expanding,li2024llava} and their medical variants~\cite{li2023llava,wang2023huatuo,xu2025lingshu}, recent work has advanced ultrasound-specific VLMs, including Dolphin~\cite{weng2025dolphin}, U2-Bench~\cite{le2025u2}, Sonomate~\cite{guo2026sonomate}, and EchoVLM~\cite{she2025echovlm}. Despite this progress, ultrasound remains challenging due to domain variability~\cite{jiang2023robotic}, speckle noise~\cite{li2025speckle2self}, and intrinsic subjectivity. Therefore, diagnosis is lesion-centric: sonographers localize suspicious regions to evaluate fine-grained attributes (e.g., echogenicity), and their judgments may vary due to inherent ambiguity.

These characteristics suggest two key requirements for ultrasound VLMs: lesion-centric structured reasoning and ambiguity-aware decision making. However, current VLMs lack explicit lesion-centric reasoning, preventing the active zooming necessary for precise diagnosis. Meanwhile, model self-consistency across multiple stochastic sampling does not naturally reflect the diagnostic ambiguity observed among sonographers (Fig.~\ref{fig:example_image}).

Recent multimodal reasoning approaches extend LLM reasoning into the visual domain~\cite{wei2022chain,wang2022self,zelikman2022star}. Grounding-based methods such as GRIT~\cite{fan2025grit} and reinforcement learning driven visual exploration models including DeepEyes~\cite{zheng2025deepeyes}, Pixel Reasoner~\cite{wang2025pixel}, and Chain-of-Focus~\cite{zhang2025chain} enable localized reasoning through bounding boxes and zooming operations. However, these methods are designed for open-world exploration with largely objective annotations typical of natural images. 
In contrast, ultrasound diagnosis is inherently ambiguous due to inter-observer variability, where uncertainty-aware reasoning is crucial. 

Uncertainty-aware VLMs have been extensively studied in the context of large language models~\cite{wang2023calibration} through confidence calibration, utilizing both black-box prompting~\cite{xiong2023can,zhou2025steerconf} and white-box logit analysis~\cite{huang2023look,kadavath2022language}. Existing VLM uncertainty research~\cite{yin2023large,stangel2025rewarding} primarily focuses on calibration against ground-truth correctness via explicitly predicting confidence (i.e., high confidence implies high accuracy, and vice versa). However, these approaches calibrate confidence against answer correctness rather than sonographer-level confidence. In the ultrasound domain, where annotations inherently reflect inter-observer variability, confidence alignment should consider sonographer uncertainty.

Taken together, multimodal reasoning and uncertainty-aware VLMs remain disconnected. Reinforcement learning can enhance reasoning by encouraging structured exploration, while multiple stochastic sampling naturally reveal answer variability reflecting model confidence. These properties make Group Relative Policy Optimization~\cite{guo2025deepseek} a natural choice, as its group-wise rollouts simultaneously strengthen reasoning and enable confidence estimation through intra-group consistency. However, prior medical GRPO-based methods~\cite{pan2025medvlm,liu2025fleming,rui2025improving} mainly use grouping to comparatively optimize accuracy-based rewards, without considering the noisy annotation and model uncertainty. 

To tackle the aforementioned limitations, we propose a sonogrpaher-inspired framework that integrates lesion-focused reasoning with an uncertainty-aware reward. First, we introduce a Zoom-then-Diagnose paradigm via supervised fine-tuning, explicitly modeling the clinical act of actively localizing and zooming into lesions before diagnosis. This paradigm improves interpretability by exposing lesion-centered reasoning steps, demonstrates genuine visual understanding, and enhances diagnostic accuracy. Second, we leverage GRPO’s group sampling mechanism to shift the reward objective from pure answer accuracy to uncertainty alignment. By treating group-wise rollout consistency as an intrinsic confidence signal and aligning it with sonographer consensus, our framework mitigates overconfidence in 
ambiguous scenarios while preserving diagnostic performance.

Our main contributions are: 
1) We introduce the first structured \textit{Zoom-then-Diagnose} paradigm, enabling lesion-focused interactive reasoning aligned with sonographer's behavior in ultrasound domain,
2) To account for ambiguity, we proposed a uncertainty-aware reward leveraging GRPO’s group-level consistency as a proxy for model confidence,
3) We demonstrate clinical validation on liver, breast, and thyroid datasets, significantly improving diagnostic accuracy and alignment with sonographer preferences in both in- and cross-domain settings.

\section{Method}
\subsection{Zoom-then-Diagnose Reasoning Dataset}
\label{sec:data}
To enable lesion-centric interactive reasoning with calibrated certainty, we build a dataset with spatial and certainty supervision. Each sample is a tuple $(I, \mathbf{b}, y, c)$, where $I$ is input image, $\mathbf{b}$ is lesion box, $y$ is diagnosis, and confidence label $c \in \{0, 1\}$ denotes consensus ($c=1$) or disagreement ($c=0$). 
We employ Qwen2.5-VL-72B [1] to structure two-stage rationales: first localize the lesion $\mathbf{b}$ (via a special \mytag{<tool\_call>} token) to obtain the cropped view $I_{\text{crop}}$, and then infer the diagnosis $y$. Rationale tone is further conditioned on 
the confidence label $c$, using definitive language for consensus (e.g., ``high certainty'') and cautious expressions for ambiguous (e.g., ``somewhat''). An example ($c=1$) follows:
\begin{tcolorbox}
[
    colback=gray!3!white,
    colframe=gray!75!black,
    boxrule=0.3pt,
    left=1mm,
    right=1mm,
    top=1mm,
    bottom=1mm,
    boxsep=1mm
]
\small
\textbf{User:} \mytag{<image>} Analyze the echo feature of the ultrasound image. \hfill $\leftarrow I$ \\[1mm]
\textbf{Assistant:} 
\mytag{<think>} Scanning the ultrasound image... I need to focus on the lesion at coordinates that match the region of interest. \mytag{</think>} \\
\mytag{<tool\_call>} \texttt{\{"bbox\_2d":{[x1,y1,x2,y2]}\}} \mytag{</tool\_call>} \hfill $\leftarrow {\mathbf{b}}$ \\[1mm]
\textbf{User:} \mytag{<image>} (Zoomed view)  \hfill $\leftarrow I_{\text{crop}}$ \\[1mm]
\textbf{Assistant:} 
\mytag{<think>} The lesion appears darker than the surrounding tissue and should be classified as hypoechoic, with \textbf{high certainty}. \mytag{</think>} \hfill $\leftarrow c$ 
\mytag{<answer>} \texttt{\{"echo": "Hypoechoic"\}} \mytag{</answer>} \hfill $\leftarrow y$
\end{tcolorbox}

\begin{figure}[h]
    \centering
    \includegraphics[width=1\textwidth]{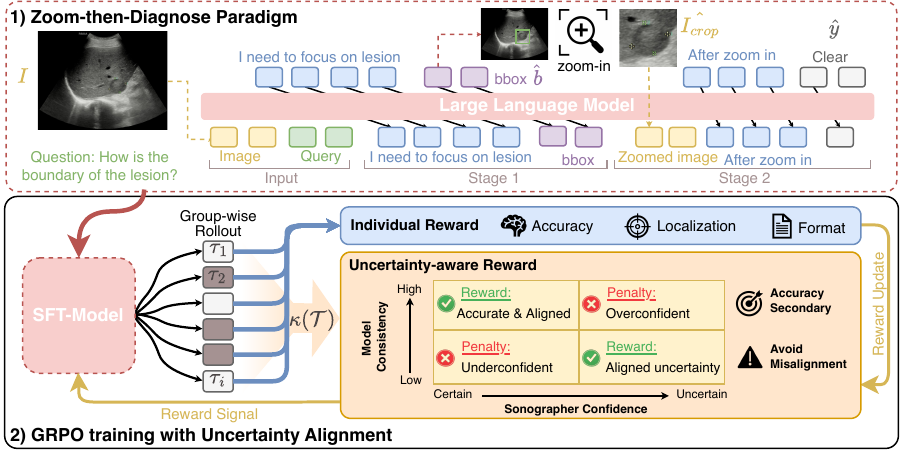}
    \caption{\textbf{Overview of the framework.} Top: The Zoom-then-Diagnose paradigm, realized by constructing a supporting dataset (Sec. 2.1) and performing supervised fine-tuning to instill structured zoom-then-diagnosis reasoning (Sec. 2.2). Bottom: We apply GRPO with a consistency-based uncertainty alignment reward (Sec. 2.3). }
    \label{fig:pipeline}
\end{figure}

\subsection{Supervised Finetuning with Zoom-then-Diagnose Reasoning}
\label{sec:sft}  

As illustrated in the upper part of Fig. \ref{fig:pipeline}, the model follows a Zoom-then-Diagnose paradigm, treating diagnosis as an interactive process. Given an ultrasound image $I$ and a textual query, the LLM first analyzes global features to generate a search rationale (e.g., “focus on the lesion”) and emits a structured \mytag{<tool\_call>} special token with predicted coordinates $\hat{\mathbf{b}}$. Rather than serving as plain text, this token functions as an executable interface that invokes a deterministic cropping function. The resulting zoomed-in region is encoded as additional visual tokens and appended to the dialogue context. Conditioned on both global and local evidence, the model then performs fine-grained reasoning to produce the final diagnosis $\hat{y}$. This Zoom–then–Diagnosis workflow is explicitly learned during supervised fine-tuning. 

\noindent \textbf{Training Details.} 
We finetune Qwen2.5-VL-7B~\cite{bai2025qwen2} using LoRA~\cite{hu2022lora} with a rank of $r=32$ . 
We explicitly optimize the language model and the alignment projector, while freezing vision tower to preserve its pretrained visual representations.

\subsection{Reinforcement Learning with Uncertainty-aware Reward}
\label{sec:rl}

\noindent \textbf{Rollout Formulation.} 
As illustrated in the bottom part of Fig.~\ref{fig:pipeline}, we define a rollout $\tau_i$ as the entire trajectory of generated tokens within a \textit{Zoom-then-Diagnose} session. Specifically, $\tau_i$ encompasses the search rationale, the predicted lesion box $\hat{\mathbf{b}_i}$, the diagnostic reasoning on the crop, and the final prediction $\hat{y}$. 
For each sample, we sample $G$ independent rollouts to form a group $\mathcal{T} = \{\tau_1, \dots, \tau_G\}$. 
The consistency across group trajectories serves as a certainty proxy $\kappa(\mathcal{T})$, aligning model hesitation with sonographer confidence $c_i$.

\noindent \textbf{Uncertainty Alignment via Group Consistency.} 
Given a set of $G$ stochastic rollouts $\mathcal{T} = \{\tau_1, \dots, \tau_G\}$, we employ Qwen2.5-VL-72B~\cite{bai2025qwen2} to parse the core diagnostic attribute $\hat{y}_i$ from each trajectory. 
We define the consensus prediction $\hat{y}^{\text{cons}}_i$ as the mode of these extracted outcomes $\{\hat{y}_i\}_{i=1}^{G}$. 
To quantify the model's internal certainty, we calculate the confidence score $\kappa(\mathcal{T})$, defined as the empirical probability of the consensus:
\begin{equation}
    \kappa(\mathcal{T}) = \frac{1}{G} \sum_{i=1}^{G} \mathbbm{1}[\hat{y}_i = \hat{y}^{\text{cons}}_i].
\label{confidence_score}
\end{equation}
where $\mathbbm{1}[\cdot]$ is the indicator function. Additionally, we define group correctness $\xi(\mathcal{T}) = \mathbbm{1}[\hat{y}^{\text{cons}}_i = y_i]$ to indicate whether the majority decision matches $y$.


We premise that for clinically certain cases ($c=1$), the model should exhibit high consistency correlated with accuracy, whereas for ambiguous cases ($c=0$), it should manifest lower agreement to reflect uncertainty. 
Accordingly, the group-level alignment reward is formulated as:
\begin{equation}
    \mathcal{R}_{\text{group}}(\mathcal{T}, c, y) = 
    \begin{cases} 
        \mathbbm{1}[\kappa(\mathcal{T}) \ge \delta] \cdot \xi(\mathcal{T}), & \text{if } c = 1 \\
        \mathbbm{1}[\kappa(\mathcal{T}) < \delta] , & \text{if } c = 0
    \end{cases}
\end{equation}
where $\delta$ is the confidence threshold. 
This formulation creates a \textit{push-pull} optimization landscape: 
for certain inputs, it rewards consistency strictly conditional on accuracy. Conversely, for ambiguous inputs, it incentivizes answer variance and forcing the model to acknowledge uncertainty instead of accuracy.

\noindent \textbf{Total Reward Composition.} 
To ensure each diagnosis is both accurate and well-aligned, we define a composite reward function for each rollout $\tau_i$. This objective integrates individual metrics with a group-level alignment update term:
\begin{equation}
    R(\tau_i) = 
    \underbrace{
        \lambda_{\text{loc}} r_{\text{loc}} 
        + \mathbbm{1}[c=1] \lambda_{\text{acc}} r_{\text{acc}} 
        + \lambda_{\text{fmt}} r_{\text{fmt}}
    }_{\text{Individual Reward}}
    + 
    \underbrace{
        \lambda_{\text{align}} \mathcal{R}_{\text{group}}(\mathcal{T}, c, y)
    }_{\text{Uncertainty Alignment}}.
\end{equation}

Here, $r_{\text{loc}}$ and $r_{\text{fmt}}$ represent reward localization IoU and valid syntax, respectively. 
Crucially, we restrict the accuracy reward $r_{\text{acc}}$ to consensus samples ($c=1$), avoiding forced certainty on ambiguous inputs.

\noindent \textbf{Training Details.}
We optimize the policy using GRPO~\cite{guo2025deepseek}, normalizing rewards within each rollout group to compute relative advantages.  For each input, $G=8$ rollouts are sampled with temperature $T=0.7$ to allow controlled exploration. 
Group consistency uses a threshold of $\delta=0.75$. 
Reward weights are set to $\lambda_{\text{loc}}=0.1$, $\lambda_{\text{acc}}=0.3$, $\lambda_{\text{fmt}}=0.1$, and $\lambda_{\text{align}}=0.5$.
\section{Experiment}

\begin{table*}[t] 
  \centering
  \caption{\textbf{Quantitative comparison.} \textbf{Acc} is evaluated on consensus samples and \textbf{mIoU} measures localization. The \colorbox{blue!6}{blue} row indicates zero-shot transfer setting.}
  \label{tab:acc}

  \resizebox{\textwidth}{!}{%
    \begin{tabular}{lcccccccccccc}
    \toprule
    & \multicolumn{2}{c}{Qwen2.5VL }
    & \multicolumn{2}{c}{Lingshu~ }
    & \multicolumn{2}{c}{MedVLM-R1 }
    & \multicolumn{2}{c}{CoF~}
    & \multicolumn{2}{c}{Qwen2.5VL$^\dagger$ }
    & \multicolumn{2}{c}{Ours} \\
    
    \cmidrule(lr){2-3}
    \cmidrule(lr){4-5}
    \cmidrule(lr){6-7}
    \cmidrule(lr){8-9}
    \cmidrule(lr){10-11}
    \cmidrule(lr){12-13}
    
    Dataset
    & Acc & mIoU
    & Acc & mIoU
    & Acc & mIoU
    & Acc & mIoU
    & Acc & mIoU
    & Acc & mIoU \\
    
    \midrule
    
    Liver
    & 47.7 & -
    & 51.1 & -
    & 53.4 & -
    & 56.8 & 13.2
    & 51.1 & 14.7
    & \textbf{69.3} & \textbf{84.7} \\
    
    Breast
    & 63.1 & -
    & 57.1 & -
    & 85.7& -
    & 65.5 & 25.9
    & 73.8 & 45.8
    & \textbf{86.9} & \textbf{71.2} \\
    
    \midrule
    
    \rowcolor{blue!6}
    Thyroid
    & 61.2 & -
    & 42.4 & -
    & \textbf{71.8} & -
    & 62.4 & 26.1
    & 69.4 & 37.4
    &  \textbf{71.8} & \textbf{60.0} \\
    
    \bottomrule
  
    \end{tabular}%
  }
\end{table*}
\textbf{Datasets.} We evaluate our framework on the public Breast Chain-of-Thought~\cite{yu2026chain} and Thyroid Nodule Ultrasound~\cite{GONG2023106389} datasets, together with an in-house Liver ultrasound dataset (205 patients with lesion-centric annotations including echogenicity, shape, and boundary). The Breast and Liver datasets are used for training (8,091 and 613 samples, respectively) and in-domain testing (120 samples per modality), while the Thyroid dataset serves as a cross-domain benchmark (120 samples) to assess generalization. This design aligns with our lesion-level reasoning focus, as attributes such as echogenicity and margins are shared across organs. Crucially, we assign binary confidence labels $c$ to distinguish between diagnostic consensus and ambiguity. For the Breast training set, confidence is derived from existing metadata (e.g., ``somewhat clear'' labeled as \textit{unconfident}). For the Liver dataset and all test sets, confidence is determined by two sonographers: diagnostic consensus is labeled as \textit{confident}, and disagreement as \textit{unconfident}.

\noindent \textbf{Baselines.} 
To ensure a fair comparison, all competing methods are based on Qwen2.5-VL-7B~\cite{bai2025qwen2}. 
We compare against: 
Qwen2.5-VL~\cite{bai2025qwen2} and Lingshu~\cite{xu2025lingshu}, representing vanilla and medically pre-trained single-turn dialog models; 
MedVLM-R1~\cite{pan2025medvlm} fine-tuned on the ultrasound data;  
Chain-of-Focus~\cite{zhang2025chain} and Qwen2.5-VL$^\dagger$~\cite{bai2025qwen2} incorporated basic tool-calling or two-turn prompting strategies, respectively, yet lack structured uncertainty-aware reward.

\subsection{Accuracy-based Evaluation}
\label{sec:experiemnt-acc}
We report Diagnostic Accuracy (Acc) and Mean Intersection over Union (mIoU) using greedy generation to assess the model's primary performance on consensus samples. However, mIoU is omitted for single-turn baselines (Qwen2.5-VL, Lingshu, MedVLM-R1) as they lack the tool-calling capability required for zoom-in.

As shown in Tab.~\ref{tab:acc}, our framework achieves the best overall performance, outperforming most baselines in both diagnostic accuracy and localization. First, despite extensive medical pre-training, Lingshu~\cite{xu2025lingshu} underperforms on lesion-specific queries, indicating that general medical knowledge cannot replace the fine-grained visual perception required for interpreting ultrasound features. 
Second, the general performance gain of CoF~\cite{zhang2025chain} and Qwen2.5VL$^\dagger$~\cite{bai2025qwen2} (with tool-call) over one-turn methods validates the necessity of a "zoom-then-diagnose" mechanism. A single pass may mix lesion features with surrounding tissues and overlook subtle lesion-specific cues (e.g., confusing a hypoechoic artery with a hypoechoic lesion), whereas a second turn explicitly restricts attention to the localized lesion region, enabling decisions based on lesion-level evidence rather than the whole image.
Third, although MedVLM-R1~\cite{pan2025medvlm} achieves the highest accuracy on the thyroid dataset, its sub-optimal performance on other organs highlights the limitations of accuracy-only rewards in subjective and noisy ultrasound settings. Further analysis is provided in the following section.

\subsection{Uncertainty-Alignment Evaluation}
\label{sec:experiemnt-align}
\begin{table}[t]
\centering
\caption{Quantitative comparison across Liver, Breast and Thyroid dataset. All values are in \%, except ECE and Gap. The \colorbox{blue!6}{blue} column indicates zero-shot transfer setting.}
\label{tab:uncertainty}

\setlength{\tabcolsep}{1.2pt}

\resizebox{\textwidth}{!}{%
\begin{tabular}{l|cccc|cccc||
>{\columncolor{blue!6}}c
>{\columncolor{blue!6}}c
>{\columncolor{blue!6}}c
>{\columncolor{blue!6}}c}
\toprule
\multirow{2}{*}{\textbf{Method}} 
& \multicolumn{4}{c|}{\textbf{Liver}} 
& \multicolumn{4}{c||}{\textbf{Breast}} 
& \multicolumn{4}{c}{\cellcolor{blue!6}\textbf{Thyroid}} \\
\cmidrule(r){2-5} \cmidrule(lr){6-9} \cmidrule(l){10-13}

& SAcc$\uparrow$ & Align$\uparrow$ & ECE$\downarrow$ & Gap$\uparrow$
& SAcc$\uparrow$ & Align$\uparrow$ & ECE$\downarrow$ & Gap$\uparrow$
& SAcc$\uparrow$ & Align$\uparrow$ & ECE$\downarrow$ & Gap$\uparrow$ \\
\midrule

Qwen2.5VL 
& 48.3 & 47.0 & 0.34 & -0.05
& 77.5 & 52.0 & 0.13 & -0.03
& 59.7 & 56.5 & 0.26 & 0.07 \\

Lingshu 
& 54.5 & 50.0 & 0.45 & 0.02
& 72.6 & 50.0 & 0.27 & 0.02
& 62.4 & 50.0 & 0.38 & 0.00 \\

MedVLM-R1 
& 55.6 & 55.4 & 0.40 & -0.03
& 85.7 & 50.0 & 0.14 & -0.02
& 70.2 & 49.4 & 0.30 & 0.02 \\

Qwen2.5VL$\dagger$
& 54.6 & 48.4 & 0.35 & -0.01
& 73.6 & 47.0 & 0.18 & 0.07
& 67.6 & 49.2 & 0.26 & 0.02 \\

CoF
& 55.2 & 55.7 & 0.44 & \textbf{0.20}
& 67.6 & 52.2 & 0.32 & 0.09
& 60.7 & 52.3 & 0.37 & 0.02 \\

\midrule

\textbf{Ours} 
& \textbf{70.8} & \textbf{67.9} & \textbf{0.13} & 0.15
& \textbf{88.9} & \textbf{69.2} & \textbf{0.09} & \textbf{0.28}
& \textbf{84.4} & \textbf{83.4} & \textbf{0.14} & \textbf{0.34} \\

\bottomrule
\end{tabular}%
}
\end{table}

We perform 8 stochastic rollouts per input with $T=0.7$ (consistent with training) to obtain the consensus prediction $\hat{y}_{\text{cons}}$ and confidence $\hat{c}$ (Eq.~\ref{confidence_score}). These are evaluated against the ground truth $y$ and confidence label $c$ using four metrics:

\noindent 
(1) Selection Accuracy: $\text{SAcc} = \frac{1}{|\mathcal{S}|} \sum_{i \in \mathcal{S}} \mathbbm{1}(\hat{y}^{\text{cons}}_i= y_i)$, where $\mathcal{S}=\{i \mid \hat{c_i} \ge \delta\}$;

\noindent 
(2) Alignment Score: $\text{Align} = \frac{1}{N} \sum_{i} \mathbbm{1} [\mathbbm{1}[\hat{c}_i \ge \delta]= c_i]$;

\noindent 
(3) Expected Calibration Error: $\text{ECE} = \sum_{m=1}^{M} \frac{|B_m|}{N} | \text{acc}(B_m) - \text{conf}(B_m) |$, where samples are grouped into $M$ confidence bins. It calculates weighted average difference between the observed accuracy and mean confidence within each group.

\noindent 
(4) Entropy Gap: $\text{Gap} = \bar{H}_{\{c=0\}} - \bar{H}_{\{c=1\}}$. This metric compares the average predictive entropy across subsets, expecting high diversity on ambiguous samples ($c_i=0$) versus consistency on consensus cases ($c_i=1$).

\begin{figure*}[t]
    \centering
    \includegraphics[width=\linewidth]{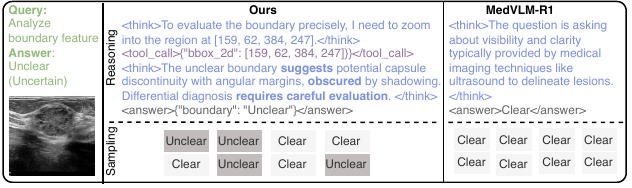}
    \caption{\textbf{Qualitative result.} Unlike MedVLM-R1's uniform predictions, our model reflects clinical ambiguity via explicit reasoning and inconsistent sampling results.}
    \label{fig:qualitative}
\end{figure*}

\textbf{Quantitative Analysis.} 
As shown in Tab.~\ref{tab:uncertainty}, our method achieves state-of-the-art performance across both seen (Liver, Breast) and unseen (Thyroid) domains. First, our model achieves superior reliability on high-confidence predictions and aligns more closely with clinical consensus, particularly in zero-shot scenarios. On the unseen Thyroid domain, it maintains the highest Selection Accuracy (82.1\%) and Alignment Score (83.4\%), significantly outperforming all baselines. These results prove our paradigm both enhances precision on confident samples and captures uncertainty, while maintaining the inherent generalization of VLMs. Second, standard VLMs (e.g., Lingshu, Qwen2.5VL) and RL-based methods like MedVLM-R1~\cite{pan2025medvlm} exhibit high ECE and low Alignment scores. This indicates a tendency towards overconfidence, while our model maintains much lower ECE (0.09 on Breast, 0.14 on Thyroid). As shown in Fig.~\ref{fig:qualitative}, unlike MedVLM-R1 which reasons with generic knowledge, our model captures lesion-specific features through active localization. By zooming into the lesion and identifying ambiguous boundary cues, it adopts a cautious stance that reflects diagnostic uncertainty. Notably, under repeated sampling, MedVLM-R1 remains consistently confident on this ambiguous case, whereas our model shows calibrated variability across rollouts, better aligning with the subjective nature of ultrasound interpretation. 
Third, many baselines exhibit near-zero or negative Entropy Gap values (e.g., Qwen2.5VL on Liver: -0.05), suggesting that their predictions remain overly consistent on clinician-labeled ambiguous cases instead of expressing higher uncertainty. In sharp contrast, our method consistently maintains a high positive Gap (0.28 on Breast, 0.34 on Thyroid), demonstrating its ability to \textit{reflect ambiguity} by mirroring sonographer disagreement on hard samples. 

\subsection{Ablation Study}
We investigate the contribution of our components on the Liver dataset by comparing SFT (no RL), accuracy-only RL, and uncertainty-based RL under the same metrics defined in Sec.~\ref{sec:experiemnt-acc} and Sec.~\ref{sec:experiemnt-align}. SFT achieves solid baseline performance in diagnostic and localization metrics (Acc 68.2\%, mIoU 85.1\%), but shows limited uncertainty alignment (SAcc 67.9\%, Align 55.10\%, Gap -0.01). The accuracy-only RL variant provides no improvement in primary performance (Acc 68.2\%, mIoU 84.3\%) and even reduces confident-sample reliability, with only marginal gains in uncertainty metrics (SAcc 65.8\%, Align 56.10\%, Gap 0.02). In contrast, our uncertainty-based RL improves both primary performance (Acc 69.3\%, mIoU 84.7\%) and uncertainty-related metrics (SAcc 70.8\%, Align 67.90\%, Gap 0.15). These results demonstrate that optimizing group-level consistency enhances both diagnostic accuracy and uncertainty calibration, whereas accuracy-only rewards provide limited benefit.










\section{Conclusion}
We consider two core challenges in ultrasound diagnosis: lesion-centric reasoning and sonographer subjectivity. We propose a Zoom-then-Diagnose paradigm that localizes the lesion via tool interaction before lesion-specific prediction, aligning model behavior with sonographer workflow. By integrating clinician confidence into training, the model reflects diagnostic ambiguity rather than producing uniformly certain outputs. While conducted on a moderately sized dataset with dual-sonographer supervision, this study establishes a scalable framework for uncertainty-aware lesion-centric reasoning. Future work will involve more sonographers and larger datasets to further enhance ultrasound image understanding.

%

%

\bibliographystyle{splncs04}

\bibliography{reference}

@article{bai2025qwen2,
  title={Qwen2. 5-vl technical report},
  author={Bai, Shuai and Chen, Keqin and Liu, Xuejing and Wang, Jialin and Ge, Wenbin and Song, Sibo and Dang, Kai and Wang, Peng and Wang, Shijie and Tang, Jun and others},
  journal={arXiv preprint arXiv:2502.13923},
  year={2025}
}

@article{li2024llava,
  title={Llava-onevision: Easy visual task transfer},
  author={Li, Bo and Zhang, Yuanhan and Guo, Dong and Zhang, Renrui and Li, Feng and Zhang, Hao and Zhang, Kaichen and Zhang, Peiyuan and Li, Yanwei and Liu, Ziwei and others},
  journal={arXiv preprint arXiv:2408.03326},
  year={2024}
}

@article{chen2024expanding,
  title={Expanding performance boundaries of open-source multimodal models with model, data, and test-time scaling},
  author={Chen, Zhe and Wang, Weiyun and Cao, Yue and Liu, Yangzhou and Gao, Zhangwei and Cui, Erfei and Zhu, Jinguo and Ye, Shenglong and Tian, Hao and Liu, Zhaoyang and others},
  journal={arXiv preprint arXiv:2412.05271},
  year={2024}
}

@article{li2023llava,
  title={Llava-med: Training a large language-and-vision assistant for biomedicine in one day},
  author={Li, Chunyuan and Wong, Cliff and Zhang, Sheng and Usuyama, Naoto and Liu, Haotian and Yang, Jianwei and Naumann, Tristan and Poon, Hoifung and Gao, Jianfeng},
  journal={Advances in Neural Information Processing Systems},
  volume={36},
  pages={28541--28564},
  year={2023}
}

@article{wang2023huatuo,
  title={Huatuo: Tuning llama model with chinese medical knowledge},
  author={Wang, Haochun and Liu, Chi and Xi, Nuwa and Qiang, Zewen and Zhao, Sendong and Qin, Bing and Liu, Ting},
  journal={arXiv preprint arXiv:2304.06975},
  year={2023}
}

@article{xu2025lingshu,
  title={Lingshu: A Generalist Foundation Model for Unified Multimodal Medical Understanding and Reasoning},
  author={Xu, Weiwen and Chan, Hou Pong and Li, Long and Aljunied, Mahani and Yuan, Ruifeng and Wang, Jianyu and Xiao, Chenghao and Chen, Guizhen and Liu, Chaoqun and Li, Zhaodonghui and others},
  journal={arXiv preprint arXiv:2506.07044},
  year={2025}
}

@article{zheng2025deepeyes,
  title={DeepEyes: Incentivizing" Thinking with Images" via Reinforcement Learning},
  author={Zheng, Ziwei and Yang, Michael and Hong, Jack and Zhao, Chenxiao and Xu, Guohai and Yang, Le and Shen, Chao and Yu, Xing},
  journal={arXiv preprint arXiv:2505.14362},
  year={2025}
}

@article{zhang2025chain,
  title={Chain-of-Focus: Adaptive Visual Search and Zooming for Multimodal Reasoning via RL},
  author={Zhang, Xintong and Gao, Zhi and Zhang, Bofei and Li, Pengxiang and Zhang, Xiaowen and Liu, Yang and Yuan, Tao and Wu, Yuwei and Jia, Yunde and Zhu, Song-Chun and others},
  journal={arXiv preprint arXiv:2505.15436},
  year={2025}
}

@article{hu2022lora,
  title={Lora: Low-rank adaptation of large language models.},
  author={Hu, Edward J and Shen, Yelong and Wallis, Phillip and Allen-Zhu, Zeyuan and Li, Yuanzhi and Wang, Shean and Wang, Lu and Chen, Weizhu and others},
  journal={ICLR},
  volume={1},
  number={2},
  pages={3},
  year={2022}
}

@article{pan2025medvlm,
  title={MedVLM-R1: Incentivizing Medical Reasoning Capability of Vision-Language Models (VLMs) via Reinforcement Learning},
  author={Pan, Jiazhen and Liu, Che and Wu, Junde and Liu, Fenglin and Zhu, Jiayuan and Li, Hongwei Bran and Chen, Chen and Ouyang, Cheng and Rueckert, Daniel},
  journal={arXiv preprint arXiv:2502.19634},
  year={2025}
}

@article{she2025echovlm,
  title={Echovlm: Dynamic mixture-of-experts vision-language model for universal ultrasound intelligence},
  author={She, Chaoyin and Lu, Ruifang and Chen, Lida and Wang, Wei and Huang, Qinghua},
  journal={arXiv preprint arXiv:2509.14977},
  year={2025}
}

@article{yu2026chain,
  title={A chain-of-thought reasoning breast ultrasound dataset covering all histopathology categories},
  author={Yu, Haojun and Li, Youcheng and Niu, Zihan and Zhang, Nan and Gong, Xuantong and Li, Huan and Zou, Zhiying and Qi, Haifeng and Cao, Zhenxiao and Lan, Zijie and others},
  journal={Scientific Data},
  year={2026},
  publisher={Nature Publishing Group}
}

@article{GONG2023106389,
title = {Thyroid region prior guided attention for ultrasound segmentation of thyroid nodules},
journal = {Computers in Biology and Medicine},
volume = {155},
pages = {106389},
year = {2023},
issn = {0010-4825},
doi = {https://doi.org/10.1016/j.compbiomed.2022.106389},
url = {https://www.sciencedirect.com/science/article/pii/S0010482522010976},
author = {Haifan Gong and Jiaxin Chen and Guanqi Chen and Haofeng Li and Guanbin Li and Fei Chen}
}

@article{guo2025deepseek,
  title={Deepseek-r1: Incentivizing reasoning capability in llms via reinforcement learning},
  author={Guo, Daya and Yang, Dejian and Zhang, Haowei and Song, Junxiao and Zhang, Ruoyu and Xu, Runxin and Zhu, Qihao and Ma, Shirong and Wang, Peiyi and Bi, Xiao and others},
  journal={arXiv preprint arXiv:2501.12948},
  year={2025}
}

@article{stangel2025rewarding,
  title={Rewarding Doubt: A Reinforcement Learning Approach to Calibrated Confidence Expression of Large Language Models},
  author={Stangel, Paul and Bani-Harouni, David and Pellegrini, Chantal and {\"O}zsoy, Ege and Zaripova, Kamilia and Keicher, Matthias and Navab, Nassir},
  journal={arXiv preprint arXiv:2503.02623},
  year={2025}
}

@article{le2025u2,
  title={U2-BENCH: Benchmarking Large Vision-Language Models on Ultrasound Understanding},
  author={Le, Anjie and Liu, Henan and Wang, Yue and Liu, Zhenyu and Zhu, Rongkun and Weng, Taohan and Yu, Jinze and Wang, Boyang and Wu, Yalun and Yan, Kaiwen and others},
  journal={arXiv preprint arXiv:2505.17779},
  year={2025}
}

@article{wang2025pixel,
  title={Pixel reasoner: Incentivizing pixel-space reasoning with curiosity-driven reinforcement learning},
  author={Wang, Haozhe and Su, Alex and Ren, Weiming and Lin, Fangzhen and Chen, Wenhu},
  journal={arXiv preprint arXiv:2505.15966},
  year={2025}
}

@article{wei2022chain,
  title={Chain-of-thought prompting elicits reasoning in large language models},
  author={Wei, Jason and Wang, Xuezhi and Schuurmans, Dale and Bosma, Maarten and Xia, Fei and Chi, Ed and Le, Quoc V and Zhou, Denny and others},
  journal={Advances in neural information processing systems},
  volume={35},
  pages={24824--24837},
  year={2022}
}

@article{wang2022self,
  title={Self-consistency improves chain of thought reasoning in language models},
  author={Wang, Xuezhi and Wei, Jason and Schuurmans, Dale and Le, Quoc and Chi, Ed and Narang, Sharan and Chowdhery, Aakanksha and Zhou, Denny},
  journal={arXiv preprint arXiv:2203.11171},
  year={2022}
}

@article{zelikman2022star,
  title={Star: Bootstrapping reasoning with reasoning},
  author={Zelikman, Eric and Wu, Yuhuai and Mu, Jesse and Goodman, Noah},
  journal={Advances in Neural Information Processing Systems},
  volume={35},
  pages={15476--15488},
  year={2022}
}

@article{fan2025grit,
  title={GRIT: Teaching MLLMs to Think with Images},
  author={Fan, Yue and He, Xuehai and Yang, Diji and Zheng, Kaizhi and Kuo, Ching-Chen and Zheng, Yuting and Narayanaraju, Sravana Jyothi and Guan, Xinze and Wang, Xin Eric},
  journal={arXiv preprint arXiv:2505.15879},
  year={2025}
}

@article{wang2023calibration,
  title={Calibration in deep learning: A survey of the state-of-the-art},
  author={Wang, Cheng},
  journal={arXiv preprint arXiv:2308.01222},
  year={2023}
}

@article{zhou2025steerconf,
  title={SteerConf: Steering LLMs for Confidence Elicitation},
  author={Zhou, Ziang and Jin, Tianyuan and Shi, Jieming and Li, Qing},
  journal={arXiv preprint arXiv:2503.02863},
  year={2025}
}

@article{xiong2023can,
  title={Can llms express their uncertainty? an empirical evaluation of confidence elicitation in llms},
  author={Xiong, Miao and Hu, Zhiyuan and Lu, Xinyang and Li, Yifei and Fu, Jie and He, Junxian and Hooi, Bryan},
  journal={arXiv preprint arXiv:2306.13063},
  year={2023}
}

@article{huang2023look,
  title={Look before you leap: An exploratory study of uncertainty measurement for large language models},
  author={Huang, Yuheng and Song, Jiayang and Wang, Zhijie and Zhao, Shengming and Chen, Huaming and Juefei-Xu, Felix and Ma, Lei},
  journal={arXiv preprint arXiv:2307.10236},
  year={2023}
}

@article{kadavath2022language,
  title={Language models (mostly) know what they know},
  author={Kadavath, Saurav and Conerly, Tom and Askell, Amanda and Henighan, Tom and Drain, Dawn and Perez, Ethan and Schiefer, Nicholas and Hatfield-Dodds, Zac and DasSarma, Nova and Tran-Johnson, Eli and others},
  journal={arXiv preprint arXiv:2207.05221},
  year={2022}
}

@article{li2025speckle2self,
  title={Speckle2Self: Self-supervised ultrasound speckle reduction without clean data},
  author={Li, Xuesong and Navab, Nassir and Jiang, Zhongliang},
  journal={Medical image analysis},
  pages={103755},
  year={2025},
  publisher={Elsevier}
}

@article{liu2025fleming,
  title={Fleming-r1: Toward expert-level medical reasoning via reinforcement learning},
  author={Liu, Chi and Li, Derek and Shu, Yan and Chen, Robin and Duan, Derek and Fang, Teng and Dai, Bryan},
  journal={arXiv preprint arXiv:2509.15279},
  year={2025}
}

@article{rui2025improving,
  title={Improving Medical Reasoning with Curriculum-Aware Reinforcement Learning},
  author={Rui, Shaohao and Chen, Kaitao and Ma, Weijie and Wang, Xiaosong},
  journal={arXiv preprint arXiv:2505.19213},
  year={2025}
}

@article{yin2023large,
  title={Do large language models know what they don't know?},
  author={Yin, Zhangyue and Sun, Qiushi and Guo, Qipeng and Wu, Jiawen and Qiu, Xipeng and Huang, Xuanjing},
  journal={arXiv preprint arXiv:2305.18153},
  year={2023}
}

@article{weng2025dolphin,
  title={Dolphin v1. 0 Technical Report},
  author={Weng, Taohan and Hu, Kaibing and Liu, Henan and Liu, Siya and Liu, Xiaoyang and Liu, Zhenyu and Ren, Jiren and Wang, Boyan and Wang, Boyang and Wang, Yiyu and others},
  journal={arXiv preprint arXiv:2509.25748},
  year={2025}
}

@article{guo2026sonomate,
  title   = {A visually grounded language model for fetal ultrasound understanding},
  author  = {Guo, Xiaoqing and Alsharid, Mohammad and Zhao, He and Wang, Yipei and Lander, Jayne and Papageorghiou, Aris T. and Noble, J. Alison},
  journal = {Nature Biomedical Engineering},
  year    = {2026},
  doi     = {10.1038/s41551-025-01578-3}
}

@article{jiang2023robotic,
  title={Robotic ultrasound imaging: State-of-the-art and future perspectives},
  author={Jiang, Zhongliang and Salcudean, Septimiu E and Navab, Nassir},
  journal={Medical image analysis},
  volume={89},
  pages={102878},
  year={2023},
  publisher={Elsevier}
}

\end{document}